\documentclass[conference]{IEEEtran}
\IEEEoverridecommandlockouts

\usepackage{cite}
\usepackage{amsmath,amssymb,amsfonts}
\usepackage{algorithmic}
\usepackage{graphicx}
\usepackage{textcomp}
\usepackage{xcolor}
\usepackage{booktabs}
\usepackage{multirow}
\usepackage{multicol}
\usepackage{hyperref}
\usepackage{url}
\usepackage[most]{tcolorbox} 
\def\BibTeX{{\rm B\kern-.05em{\sc i\kern-.025em b}\kern-.08em
    T\kern-.1667em\lower.7ex\hbox{E}\kern-.125emX}}

\begin{document}

\title{Ensembles of Large Language Models for Identifying EQ-5D Studies in PubMed Based on Their Abstracts}

\author{
\IEEEauthorblockN{
Zhyar Rzgar K. Rostam\IEEEauthorrefmark{1}\IEEEauthorrefmark{2},
M\'{a}rta P\'{e}ntek\IEEEauthorrefmark{6}\IEEEauthorrefmark{4},
J\'{a}nos Tibor Czere\IEEEauthorrefmark{4}\IEEEauthorrefmark{5},\\
Zsombor Zrubka\IEEEauthorrefmark{6}\IEEEauthorrefmark{4},
L\'{a}szl\'{o} Gul\'{a}csi\IEEEauthorrefmark{6}\IEEEauthorrefmark{4},
and G\'{a}bor Kert\'{e}sz\IEEEauthorrefmark{2}\IEEEauthorrefmark{3}
}

\IEEEauthorblockA{\IEEEauthorrefmark{1}Doctoral School of Applied Informatics and Applied Mathematics, Obuda University, Budapest, Hungary}
\IEEEauthorblockA{\IEEEauthorrefmark{2}John von Neumann Faculty of Informatics, Obuda University, Budapest, Hungary}
\IEEEauthorblockA{\IEEEauthorrefmark{3}Laboratory of Parallel and Distributed Systems,\\ Institute for Computer Science and Control (SZTAKI), Hungarian Research Network (HUN-REN), Budapest, Hungary}
\IEEEauthorblockA{\IEEEauthorrefmark{4}Doctoral School of Innovation Management, Obuda University, Budapest, Hungary}
\IEEEauthorblockA{\IEEEauthorrefmark{5}PSI CRO Hungary LLC, Budapest, Hungary}
\IEEEauthorblockA{\IEEEauthorrefmark{6}HECON Health Economics Research Center, University Research and Innovation Center,\\ Obuda University, Budapest, Hungary}

\IEEEauthorblockA{Emails: \{kwekha.rostam.zhyar, pentek.marta, czere.janos, zrubka.zsombor, gulacsi, kertesz.gabor\}@uni-obuda.hu}
}



\maketitle

\begin{abstract}
The rapid increase in scientific publications leads to the fact that manual study screening in systematic literature reviews (SLRs) is increasingly resource consuming, inefficient, and inconsistent. Classifying studies that clearly report health-related quality-of-life results, such as EQ-5D data, requires a high level of clinical interpretation and poses challenges for human reviewers. This study investigates the use of Google's Gemini and Gemma large language models (LLMs) in automating EQ-5D detection in the PubMed biomedical database based only on published abstracts. A multi-phase framework is proposed that integrates few-shot prompting, weight ensembling aggregation, and a soft stacking meta-classifier. Nine LLMs are evaluated on a dataset of PubMed studies manually labeled by two experts regarding EQ-5D reporting. The weighted ensemble of \texttt{gemini-2.5-pro}, \texttt{gemma-3-12b}, and \texttt{gemma-3-27b} obtained a 0.74 weighted F1-score and 0.74 accuracy, exceeding individually attained results. The ensembling of top-performing models improved the balance between precision and recall compared to individual models, while the soft stacking approach provided greater reliability and interpretability. Feature analysis shows that the probability results from the models are important in guiding the final predictions. The findings suggest that an ensemble-based LLM setup is a reliable and scalable approach for automating screening in biomedical research.

\end{abstract}

\begin{IEEEkeywords}
Biomedical Text Mining, Health-Related Quality of Life, Few-Shot Prompting, Systematic Literature Review, Natural Language Processing
\end{IEEEkeywords}

\section{Introduction}
\label{introduction}
In the past decades, systematic literature reviews (SLRs) have been utilized as a key approach for gathering and summarizing evidence for medical and financial decision making~\cite{ahanger2022novel, app13063594}. The continuous growth of scientific publications has made the manual screening process increasingly slow, inconsistent, and exposed to human error. This challenge becomes more significant when complex inclusion criteria are used, such as to identify studies that specifically report results with a widely used health-related quality of life measurement tool, the EQ-5D~\cite{kertesz2024eq5d, zah2022paying}.

The EQ-5D, developed by the EuroQol group, comprises a descriptive system covering five domains of health: \textit{mobility, self-care, usual-activities, pain/discomfort, and anxiety/depression}. Self-reported problem levels on the five domains can be converted into an index value. The second part of the EQ-5D questionnaires is a 0-100 visual analogue scale (0 represents the worst health status the respondent can imagine, and 100 represents the best health status the respondent can imagine) to assess the respondent's self-perceived global health. There are different versions of the EQ-5D (e.g., 3 and 5 level response scale in the descriptive system, youth version), but the EQ-5D index can be derived from all. The EQ-5D index is the most commonly used measure to calculate health benefits in cost-utility analysis aiming to inform health policy decisions~\cite{1990199, szende2014self, kertesz2024eq5d}. Automation of systematic literature reviews (SLRs) to gather all relevant studies presenting EQ-5D data is very important as going through a large number of papers manually to find them is time consuming and inconsistent.

These challenges highlight the demand for automated tools that support study screening and classification with high accuracy. Recent developments in the field of natural language processing (NLP)~\cite{peng2021survey, naseem2022benchmarking} and deep learning (DL)~\cite{jiao2023brief, alimova2021cross}, particularly transformer-based large language models (LLMs), have advanced automated text classifications~\cite{10.1145/3763002}. LLMs have recently been applied to automate the systematic literature review process, but the need for improving and standardizing the reporting of clinical studies is still considerable~\cite{zrubka2025artificial}. Although general-purpose pre-trained language models (PLMs) are efficient in a wide variety of text tasks, they often have difficulties in handling domain-specific texts such as biomedical contexts~\cite{beltagy2019scibert, 10.1145/3763002, lee2020biobert, peng2019transfer, liu2021finbert, kim2023medibiodeberta, 10820432}.

This study introduces an ensemble strategy built upon Google’s Gemini and Gemma LLMs to automatically detect publications reporting EQ-5D results in biomedical texts published in PubMed biomedical literature database using only publicly available abstract information. We propose a multi-phase approach\footnote{The datasets, model implementations, and examples of results can be found at this account: \url{https://github.com/ZhyarUoS/}, under the repository Ensemble-LLMs-for-Classifying-EQ-5D-Reporting-from-Abstracts.} that involves the use of few-shot prompting, weighted ensemble learning, and soft stacking to increase performance and reliability.

The principal aims of this research are to:

\begin{itemize}
    \item Assess the performance of nine Gemini and Gemma models using few-shot prompting in EQ-5D classification.
    \item Propose a weighted ensemble method that combines prediction and confidence levels.
    \item Present a soft stacking framework using probability and confidence attributes to achieve higher generalization.
    \item Provide a comparison among ensemble and stacking approaches with individual LLM.
\end{itemize}

\section{Related Works}
\subsection{LLMs for Biomedical Literature Screening}
Cao et al.~\cite{doi:10.7326/ANNALS-24-02189} introduce a framework combining Chain-of-Thought and Instruction-Structure-Optimized prompting with LLMs to perform systematic review screening at human-level or superior accuracy. GPT-4 reaches a maximum accuracy of 93.6\% and sensitivity of 97.5\% with the BenchSR benchmark, outperforming human reviewers in several cases. Their approach reduced screening time and cost by more than 90\%, advancing toward full automation of systematic reviews.

\subsection{Ensemble and Stacking Approaches}
Abburi et al.~\cite{abburi2023generative} propose an ensemble LLM framework for detecting AI-generated text and attributing it to specific language models. Their approach is based on the combination of probabilities from fine-tuned LLMs such as BERT, RoBERTa, DeBERTa, and XLM-RoBERTa with conventional machine learning classifiers. The results indicate strong cross-lingual performance on the AuTexTification benchmark, being one of the best in AI-vs-human detection.

Ridoy et al.~\cite{10825609} develop EnStack, an ensemble stacking approach that combines multiple language models (CodeBERT, GraphCodeBERT, and UniXcoder) to improve software vulnerability detection. Each model is fine-tuned on the Draper VDISC dataset, and their predictions are combined with a meta-classifier like Logistic Regression, SVM, Random Forest, and XGBoost. EnStack is able to achieve an accuracy of 82.36\% and an AUC of 92.85\% by combining semantic, structural, and cross-modal representations of code. This setup is able to outperform individual model performance and demonstrate its effectiveness for vulnerability detection.

\subsection{Domain-Specific PLMs}
Rostam and Kertész~\cite{10.1145/3763002} compared general-purpose PLM (BERT) with the domain-specific PLMs (BioBERT and SciBERT) for sentence classification in the biomedical abstracts from the PubMed 20k dataset. Their findings indicate that fine-tuned domain-specific models are more accurate than BERT. The findings show that PLMs trained on a specific domain are beneficial for facilitating better comprehension of biomedical texts.

\section{Methodology}
\subsection{Dataset Description}
\label{dataset_description}
The dataset employed in this work is prepared by Kertész et al.~\cite{kertesz2024eq5d}. In brief, the records were obtained from PubMed, a database of biomedical literature that contains more than 36 million citations~\cite{white2020pubmed}. The original dataset for the study contains all the EQ-5D publications that were identified using EuroQol Group's search tool developed for searching EQ-5D in PubMed. A total of 15,547 studies published between 1990 and 2022 were identified. A total of 200 publications were randomly selected from the output using the sample command in Stata 2017. Two independent experts fully reviewed the full text of each of the 200 selected studies. This guaranteed that the full paper reported EQ-5D results (either the EQ-5D descriptive system, the EQ-5D index score, or the EQ VAS results of any official EQ-5D version, using any administration method, such as digital or paper-based formats and self-complete or proxy versions).  In this selection process, the reviewers followed strict inclusion and exclusion criteria. The reviewers verified whether the use and results of the EQ-5D instrument were explicitly reported in the publication. Any disagreement in labeling was resolved through discussion until consensus was achieved. As a result, the extracted dataset contains 121 positive (EQ-5D results in the paper) and 79 negative (no EQ-5D results presented in the paper) publications. The abstracts of the publications were all in English, although some full texts were in other languages. These publications were labeled negative and treated as a minor source of noise.

\subsection{Dataset Preparation and Preprocessing}
Each record in the dataset (see Section~\ref{dataset_description}) included abstract text, a unique identifier, a title, keywords, and a binary label indicating whether the study explicitly reported EQ-5D data. The abstract texts are extracted and preprocessed for input to the model, and they are inserted into structured prompts, which are made for binary classification.

A subset of 40 examples, balanced by including 20 positive and 20 negative samples, is randomly selected from the dataset to construct a few-shot demonstration prompts. Each abstract is appended after a set of labeled examples within the prompt, which assisted the model in inferring the correct class based on the prior demonstrations. As a result, the models are expected to provide a textual output containing both a predicted class with a confidence level value (0-100). These responses are utilized for consecutive evaluation and ensemble analysis.

\subsection{LLM Comparison}
In this section, we present the investigated methodological pipelines for evaluating LLMs' ability to detect EQ-5D within clinical study abstracts.

\subsubsection{Few-Shot Prompting Strategy}
All experiments are conducted using a few-shot prompting setup to guide model reasoning using in-context examples. Each prompt has a task description followed by labeled demonstration examples (20 positive and 20 negative) extracted from the labeled dataset. The LLMs must determine whether the study mentioned the EQ-5D instrument versions or their variant names (e.g. EQ-5D-3L, EQ-5D-5L, EuroQol-5D) in the study by investigating only the abstract text (see Eq.~\eqref{eq:prompt}). The prompt template is stated below:

\begin{tcolorbox}[colback=gray!10, colframe=black, boxrule=0.5pt, arc=2pt]
\small
\textit{You are a biomedical text classification expert.}

\textit{Task: Determine whether the following clinical study abstract provides explicit evidence that the EQ-5D instrument (or a variant like EQ-5D-3L, EQ-5D-5L, EuroQol-5D) was actually used in the study's methods or results.}
\\

\textit{Answer format (exactly one line):}\\
Prediction: [Yes/No]\\
Confidence: [number between 0 and 100]
\\

\textit{Few-shot examples: [Insert few-shot samples]}

\textit{Now classify this new abstract:}\\
Abstract: [Insert abstract text here]
\end{tcolorbox}

Given an example set \(E=\{(A_i,y_i)\}_{i=1}^{N}\), the prompt is defined as:
\begin{equation}
\text{Prompt}(A,E)=\text{Instr.}+\sum_{i=1}^{N}\text{Ex.}(A_i,y_i)+\text{Qry.}(A)
\label{eq:prompt}
\end{equation}
Where \(A\) denotes the target abstract, \(\text{Instr.}\) is the task instruction, 
\(\text{Ex.}(A_i,y_i)\) represents the \(i^{th}\) labeled demonstration, and \(\text{Qry.}(A)\) 
is the evaluated abstract.

\subsubsection{Investigated LLMs}
Nine different models from Google's Gemini and Gemma families are evaluated independently using a few-shot setup (see Table~\ref{tab:google_models}). All models are tested on 200 manually labeled studies, with their predictions parsed and assessed using weighted F1-scores (see Eq.~\eqref{eq:weighted_f1}).

\begin{table}[h]
\centering
\caption{Details for Google's Gemini and Gemma Models}
\label{tab:google_models}
\begin{tabular}{lll}
\hline
\textbf{Model ID} & \textbf{Family} & \textbf{Size (Parameters)} \\
\hline
google/gemini-2.0-flash & Gemini 2.0 & – \\
\hline
google/gemini-2.0-flash-lite & Gemini 2.0 & – \\
\hline
google/gemini-2.5-flash & Gemini 2.5 & – \\
\hline
google/gemini-2.5-flash-lite & Gemini 2.5 & – \\
\hline
google/gemini-2.5-pro & Gemini 2.5 & – \\
\hline
google/gemma-3-12b & Gemma 3 & 12B \\
\hline
google/gemma-3-1b & Gemma 3 & 1B \\
\hline
google/gemma-3-27b & Gemma 3 & 27B \\
\hline
google/gemma-3-4b & Gemma 3 & 4B \\
\hline
\end{tabular}
\end{table}

Weighted F1 is being used to compute all sorts of performance results:

\begin{equation}
F1_{\text{weighted}} = \frac{\sum_{c \in \{0,1\}} n_c \cdot F1_c}{\sum_{c \in \{0,1\}} n_c}
\label{eq:weighted_f1}
\end{equation}

Where $n_c$ denotes the number of samples in class $c$ and $F1_c$ is the per-class F1-score.

\subsubsection{Pruned Model Selection}
After the individual evaluation process, the top three models with the best-weighted F1 scores are selected for ensemble combination:

\begin{itemize}
    \item \texttt{google/gemini-2.5-pro}
    \item \texttt{google/gemma-3-12b}
    \item \texttt{google/gemma-3-27b}
\end{itemize}
As a result, these models are used as the basis for the pruned ensemble stage, since they consistently showed excellent performance in classification accuracy and confidence calibration.

\subsubsection{Weighted Ensemble Framework}
We use the achieved weighted F1-scores for creating a weighted ensemble by combining model predictions. Every model's prediction (vote) is scaled by its F1 performance and prediction confidence. For each abstract, both classes are considered, the ensemble accumulated positive and negative scores across models based on the equations~\eqref{eq:s_yes},~\eqref{eq:s_no}, and \eqref{eq:y_hat}.

\begin{equation}
S_{\text{Yes},i} = \sum_m w_m \cdot c_{m,i} \cdot \mathbf{1}(p_{m,i} = 1)
\label{eq:s_yes}
\end{equation}

\begin{equation}
S_{\text{No},i} = \sum_m w_m \cdot c_{m,i} \cdot \mathbf{1}(p_{m,i} = 0)
\label{eq:s_no}
\end{equation}

\begin{equation}
\hat{y}_i =
\begin{cases}
1, & \text{if } S_{\text{Yes},i} \ge S_{\text{No},i} \\
0, & \text{otherwise}
\end{cases}
\label{eq:y_hat}
\end{equation}

Where:  
\begin{itemize}
    \item $p_{m,i}$ is the binary prediction from model $m$ for sample $i$,
    \item $c_{m,i}$ is the confidence of model $m$ for sample $i$, normalized to $[0,1]$,
    \item $w_m$ is the weight of model $m$, corresponding to its individual F1-score,
    \item $\mathbf{1}(\cdot)$ is the indicator function, equal to 1 if the condition is true and 0 otherwise.
\end{itemize}

\subsubsection{Soft Stacking Framework}
In addition to ensemble voting, a soft stacking model is introduced for the purpose of finding the most appropriate combinations of model predictions and confidence scores.

For each base model $m$ and sample $i$, the model's prediction $p_{m,i}$ and confidence $c_{m,i}$ are used to obtain a soft probability of predicting the positive class (see Eq.~\eqref{eq:soft_prob}):

\begin{equation}
P_{m,i} =
\begin{cases}
c_{m,i}, & \text{if } p_{m,i} = 1 \\
1 - c_{m,i}, & \text{if } p_{m,i} = 0
\end{cases}
\label{eq:soft_prob}
\end{equation}

The meta-feature vector for sample $i$ is defined as (see Eq.~\eqref{eq:meta_features}):

\begin{equation}
X_i = [P_{1,i}, P_{2,i}, P_{3,i}, c_{1,i}, c_{2,i}, c_{3,i}]
\label{eq:meta_features}
\end{equation}

The soft positive class probabilities are represented by the first three elements, whereas the raw confidence values from each model are the attributes of the last three elements.

The feature set is comprised of these elements, and the logistic regression meta-classifier is the model trained based on the predicted probabilities in 5-fold stratified cross-validation technique (see Eq.~\eqref{eq:logreg_meta}).

\begin{equation}
\hat{y}_i = \sigma(w^T X_i + b), \quad 
\sigma(z) = \frac{1}{1 + e^{-z}}
\label{eq:logreg_meta}
\end{equation}

The learned coefficients $w$ quantified the contribution of the probability and confidence of each model to the final decision of the ensemble.

\begin{table}[h]
\centering
\caption{Overview of the multi-phase ensemble framework.}
\label{tab:ensemble_phases}
\resizebox{0.45\textwidth}{!}{%
\begin{tabular}{lllll}
\hline
\textbf{Phase} & \textbf{Models} & \textbf{Approach} & \textbf{Learning} & \textbf{Metric} \\
\hline
Phase 1  & 9 Few-Shot & Independent & Prompt-based & Weighted F1 \\
Phase 2a & Top 3     & Weighted Ensemble & Confidence-voting & Weighted F1 \\
Phase 2b & Top 3     & Soft Stacking     & Logistic Regression & 5-fold CV F1 \\
\hline
\end{tabular}%
}
\end{table}

\section{Results}
\label{sec:results}
In this section, we present the results obtained by all the pipelines implemented. The evaluation aims at assessing models' performance across different experimental setups and to highlight the impact of methodological choice on the outcome.

\subsection{Individual Model Performance}
\label{sec:individual_model_perf}
This phase included the evaluation of nine LLMs on the labeled dataset (see Section~\ref{dataset_description}). The weighted F1-score was used as the primary performance metric (see Table~\ref{tab:individual_results}, all metrics except Accuracy are weighted to account for class imbalance).

\begin{table}[h]
\centering
\caption{Individual model performance on the 200-study dataset (all metrics weighted except Accuracy)}
\label{tab:individual_results}
\scalebox{0.9}{%
\begin{tabular}{lcccc}
\hline
\textbf{Model} & \textbf{Precision} & \textbf{Recall} & \textbf{Accuracy} & \textbf{F1-score} \\
\hline
Gemini-2.0-Flash        & 0.72 & 0.66 & 0.66 & 0.58 \\
Gemini-2.0-Flash-Lite   & 0.70 & 0.62 & 0.62 & 0.51 \\
Gemini-2.5-Flash        & 0.66 & 0.65 & 0.65 & 0.58 \\
Gemini-2.5-Flash-Lite   & 0.70 & 0.67 & 0.67 & 0.61 \\
Gemini-2.5-Pro          & \textbf{0.75} & \textbf{0.73} & \textbf{0.73} & \textbf{0.71} \\
Gemma-3-1B              & 0.52 & 0.56 & 0.56 & 0.52 \\
Gemma-3-4B              & 0.57 & 0.59 & 0.59 & 0.57 \\
Gemma-3-12B             & 0.70 & 0.65 & 0.65 & 0.65 \\
Gemma-3-27B             & 0.66 & 0.67 & 0.67 & 0.65 \\
\hline
\end{tabular}%
}
\end{table}


According to individual results (presented in Table~\ref{tab:individual_results}), three models: \texttt{google/gemini-2.5-pro}, \texttt{google/gemma-3-12b}, and \texttt{google/gemma-3-27b} demonstrated superior weighted F1-score (0.71, 0.65, and 0.65, respectively) among the nine evaluated LLMs. In particular, \texttt{gemini-2.5-pro} achieved the highest and most balanced performance, effectively distinguishing both positive and negative samples.

\subsection{Weighted Ensemble Performance}
\label{sec:weight_ensemble_results}
In this configuration, the top three LLMs (\texttt{google/gemini-2.5-pro}, \texttt{google/gemma-3-12b}, and \texttt{google/gemma-3-27b}) were combined together, where each model’s vote was weighted by its F1-score and prediction confidence, leading to performance improvement. The ensemble obtained a general accuracy of 0.74 and a weighted F1-score of 0.74. With the weighted aggregation of model predictions, the results demonstrated improved generalization and balanced performance for both classes, with 0.79 and 0.65 achieved for the positive and negative classes, respectively. Therefore, according to the results, the weighted aggregation of model predictions consistently outperforms the models when considered individually (see Table~\ref{tab:ensemble_improvement}).

\begin{table}[h]
\centering
\caption{Performance improvements of the pruned ensemble over the best single model}
\label{tab:ensemble_improvement}
\begin{tabular}{lccc}
\hline
\textbf{Metric} & \textbf{Ensemble} & \textbf{Best Single Model} & \textbf{Improvement} \\
\hline
Weighted Recall   & \textbf{0.74} & 0.73 (Gemini-2.5-Pro) & \textbf{+0.01} \\
Accuracy          & \textbf{0.74} & 0.73 (Gemini-2.5-Pro) & \textbf{+0.01} \\
Weighted F1-score & \textbf{0.74} & 0.71 (Gemini-2.5-Pro) & \textbf{+0.03} \\
Macro F1-score    & \textbf{0.72} & 0.69 (Gemini-2.5-Pro) & \textbf{+0.03} \\
\hline
\end{tabular}
\end{table}



\subsection{Soft Stacking Performance}
\label{sec:soft_stacking_results}
The soft stacking approach consisted of performing logistic regression on six meta-features (soft positive probabilities and raw confidence values from each base model) using 5-fold stratified cross-validation. The stacking model obtained a weighted F1-score of 0.72 and an overall accuracy of 0.73 (as shown in Table~\ref{tab:stacking_results}).

\begin{table}[h]
\centering
\caption{Performance of the soft stacking model (5-fold CV).}
\label{tab:stacking_results}
\begin{tabular}{lccc}
\hline
\textbf{Class} & \textbf{Precision} & \textbf{Recall} & \textbf{F1-score} \\
\hline
Negative & 0.73 & 0.51 & 0.60 \\
Positive & 0.73 & 0.88 & 0.80 \\
\hline
\textbf{Overall Accuracy} & \multicolumn{3}{c}{0.73} \\
\textbf{Weighted F1} & \multicolumn{3}{c}{0.72} \\
\hline
\end{tabular}
\end{table}

Feature importance analysis (Table~\ref{tab:stacking_importance}) showed soft probabilities from \texttt{google/gemini-2.5-pro}, \texttt{google/gemma-3-12b}, and \texttt{google/gemma-3-27b} to be strong features, with coefficients equal to 2.21, 1.15, and 0.75, respectively. Confidence scores gave marginal information to the final decision, suggesting that most of the discriminative signal was contained in the model probabilities for this ensemble.

\begin{table}[h]
\centering
\caption{Top feature coefficients from the soft stacking meta-model.}
\label{tab:stacking_importance}
\begin{tabular}{lc}
\hline
\textbf{Feature} & \textbf{Coefficient} \\
\hline
\texttt{gemini-2.5-pro\_prob\_yes} & 2.213 \\
\texttt{gemma-3-12b\_prob\_yes} & 1.155 \\
\texttt{gemma-3-27b\_prob\_yes} & 0.746 \\
\texttt{gemma-3-12b\_conf} & 0.253 \\
\texttt{gemini-2.5-pro\_conf} & 0.072 \\
\texttt{gemma-3-27b\_conf} & -0.030 \\
\hline
\end{tabular}
\end{table}

Overall, this approach was able to maintain a good balance between recall and precision, showing that combining different base model methods can significantly increase both performance and stability. The interpretability of the coefficients suggests that LLM-derived probabilities were reliable meta-features for biomedical text classification.

\subsection{Practical Feasibility and Cost Analysis}
The feasibility of the proposed approach for large-scale screening is evaluated by measuring the runtime and the approximate inference costs per model. Based on the size and configuration of the model, the time required to process abstracts ranges from approximately 7 to 64 minutes per experiment. The cost ranged from 0.07 to 5.04 USD per run, according to the official Gemini API estimates for Gemini models~\cite{GooglePricing2026} and the community estimates for Gemma models~\cite{PricePerToken2026} (see Table~\ref{tab:runtime_cost}).

\begin{table}[h]
\centering
\caption{Runtime and estimated cost for processing 200 abstracts.}
\label{tab:runtime_cost}
\begin{tabular}{lcc}
\hline
\textbf{Model} & \textbf{Runtime (min)} & \textbf{Cost (USD)} \\
\hline
Gemini-2.0-Flash      & 7.7  & 0.40 \\
Gemini-2.0-Flash-Lite & 10.4 & 0.30 \\
Gemini-2.5-Flash      & 15.0 & 1.21 \\
Gemini-2.5-Flash-Lite & 7.1  & 0.40 \\
Gemini-2.5-Pro        & 64.3 & 5.04 \\
Gemma-3-12B           & 19.9 & 0.56 \\
Gemma-3-1B            & 19.7 & 0.08 \\
Gemma-3-27B           & 18.8 & 0.40 \\
Gemma-3-4B            & 12.3 & 0.07 \\
\hline
\end{tabular}

\end{table}

\section{Discussion}
The results showed that LLMs can effectively detect EQ-5D reporting in clinical study abstracts, although the results were varied depending on the model used and configuration.

\subsection{Individual Model Performance Analysis}
According to the achieved results (see Section~\ref{sec:individual_model_perf}, and Table~\ref{tab:individual_results}), among the nine different LLMs, \texttt{google/gemini-2.5-pro} demonstrated the best overall performance by obtaining a weighted F1-score 0.71, and an accuracy 0.73, and followed by \texttt{gemma-3-12b} and \texttt{gemma-3-27b}, both of which achieved 0.65 for weighted F1-score. The results showed that the large models consistently outperformed smaller ones, indicating that the higher parameter capacity improved the ability to understand the biomedical context. 

\subsection{Ensemble and Stacking Performance}
The obtained results from ensembling weighted phase (see Section~\ref{sec:weight_ensemble_results}), indicated that by combining the top three models through a weighted ensemble approach, the overall performance was increased (weighted F1-score was 0.74, and accuracy was 0.74) showing a better trade-off between sensitivity and specificity. This demonstrates that combining model predictions reduces the biases of individual models and assists in generalizing the prediction.

In the soft stacking approach, we utilized logistic regression on model probabilities and confidence values. This setup was able to achieve comparable performance (weighted F1 score was 0.72; accuracy was 0.73). Feature analysis showed that soft probabilities from \texttt{gemini-2.5-pro}, \texttt{gemma-3-12b}, and \texttt{gemma-3-27b} were the most influential, confirming that probabilistic model outputs have stronger discriminative information than raw confidence scores.

\subsection{Feasibility and Resource Considerations}
The experiments show that the approach is scalable and cost-effective for large-scale screening. A high-performing model, Gemini-2.5-Pro, achieved the best accuracy and F1-score but was computationally intensive (64 min) and with the highest cost (5.04 USD). In contrast, lighter models such as Gemini-2.5-Flash-Lite and Gemma-3-4B demonstrated moderate performance and completed the task in under 13 minutes at a minimal cost of 0.07–0.40 USD (see Table~\ref{tab:runtime_cost}). This demonstrates a practical balance between accuracy, runtime, and cost. Lower cost models achieved acceptable accuracy, supporting scalable implementation in resource constrained settings.

\section{Conclusion and Future Directions}
This study proposed an ensemble-based LLM framework for the automation of the detection of EQ-5D instrument reporting in biomedical abstracts. The approach can effectively automate EQ-5D detection in biomedical abstracts in PubMed. The best performance (weighted F1-score = 0.74) was achieved by the weighted ensemble of \texttt{gemini-2.5-pro}, \texttt{gemma-3-12b}, and \texttt{gemma-3-27b} as compared with the individual models. The findings confirmed that combining model probabilities and confidence scores improves reliability, balance, and interpretability in automated screening. Future study directions can be in many directions, such as:
\begin{itemize}
    \item Enlarge the dataset by covering more medical studies with the dataset to ensure improved results.
    \item Fine-tune on several biomedical PLMs (for example, BioBERT, PubMedBERT).
    \item Explore co-training approaches to work with labeled and unlabeled data.
\end{itemize}

\section{Limitations}
This section outlines the main limitations in our study:
\begin{itemize}
    \item The dataset (200 studies) size may limit generalization to broader corpora. 
    \item Investigation was only restricted to abstracts published in the PubMed database, thus the method cannot be directly transferred to other literature databases.
    \item The study was evaluated on a single dataset, which may limit generalization.

\end{itemize}

\section{Acknowledgment}
The authors gratefully acknowledge the members of the Applied Machine Learning Research Group at Obuda University's John von Neumann Faculty of Informatics for their insightful feedback and suggestions. They also wish to acknowledge the support provided by the Doctoral School of Applied Informatics and Applied Mathematics at Obuda University. The work of MP, ZZ and GL has been supported by the National Research, Development, and Innovation Fund of Hungary, financed under the TKP2021-NKTA-36 funding scheme at Obuda University. 
\bibliographystyle{ieeetr}
\bibliography{ref}

\begin{thebibliography}{10}

\bibitem{ahanger2022novel}
M.~M. Ahanger and M.~A. Wani, ``Novel deep learning approach for scientific literature classification,'' in {\em 2022 9th International Conference on Computing for Sustainable Global Development (INDIACom)}, pp.~249--254, IEEE, 2022.

\bibitem{app13063594}
R.~Alfaro, H.~Allende-Cid, and H.~Allende, ``Multilabel text classification with label-dependent representation,'' {\em Applied Sciences}, vol.~13, no.~6, 2023.

\bibitem{kertesz2024eq5d}
G.~Kertész, J.~T. Czere, Z.~Zrubka, L.~Gulácsi, and M.~Péntek, ``Towards automating the selection of articles reporting eq-5d data for systematic literature reviews using large language models.'' \url{https://ssrn.com/abstract=4876024}, 2024.
\newblock Available at SSRN: \url{https://ssrn.com/abstract=4876024} or \url{http://dx.doi.org/10.2139/ssrn.4876024}.

\bibitem{zah2022paying}
V.~Zah, A.~Burrell, C.~Asche, and Z.~Zrubka, ``Paying for digital health interventions--what evidence is needed?,'' {\em Acta Polytechnica Hungarica}, vol.~19, no.~9, pp.~179--199, 2022.

\bibitem{1990199}
``Euroqol - a new facility for the measurement of health-related quality of life,'' {\em Health Policy}, vol.~16, no.~3, pp.~199--208, 1990.

\bibitem{szende2014self}
A.~Szende, B.~Janssen, and J.~Cabases, eds., {\em Self-Reported Population Health: An International Perspective based on EQ-5D}.
\newblock Dordrecht, Netherlands: Springer Nature, 2014.
\newblock \url{https://www.ncbi.nlm.nih.gov/pubmed/29787044}, PMID: 29787044.

\bibitem{peng2021survey}
J.~Peng and K.~Han, ``Survey of pre-trained models for natural language processing,'' in {\em 2021 International Conference on Electronic Communications, Internet of Things and Big Data (ICEIB)}, pp.~277--280, IEEE, 2021.

\bibitem{naseem2022benchmarking}
U.~Naseem, A.~G. Dunn, M.~Khushi, and J.~Kim, ``Benchmarking for biomedical natural language processing tasks with a domain specific albert,'' {\em BMC bioinformatics}, vol.~23, no.~1, p.~144, 2022.

\bibitem{jiao2023brief}
Q.~Jiao, ``A brief survey of text classification methods,'' in {\em 2023 IEEE 3rd International Conference on Information Technology, Big Data and Artificial Intelligence (ICIBA)}, vol.~3, pp.~1384--1389, IEEE, 2023.

\bibitem{alimova2021cross}
I.~Alimova, E.~Tutubalina, and S.~I. Nikolenko, ``Cross-domain limitations of neural models on biomedical relation classification,'' {\em IEEE Access}, vol.~10, pp.~1432--1439, 2021.

\bibitem{10.1145/3763002}
Z.~R. K.~Rostam and G.~Kert\'{e}sz, ``Advances in pre-trained language models for domain-specific text classification: A systematic review,'' {\em ACM Trans. Intell. Syst. Technol.}, vol.~16, Oct. 2025.

\bibitem{zrubka2025artificial}
Z.~Zrubka, L.~Kov{\'a}cs, H.~Motahari~Nezhad, J.~Czere, L.~Gul{\'a}csi, and M.~P{\'e}ntek, ``Artificial intelligence in medicine: A systematic review of guidelines for the reporting and interpretation of studies,'' {\em Acta Polytechnica Hungarica}, vol.~22, no.~10, pp.~125--143, 2025.

\bibitem{beltagy2019scibert}
I.~Beltagy, K.~Lo, and A.~Cohan, ``Scibert: A pretrained language model for scientific text,'' {\em arXiv preprint arXiv:1903.10676}, 2019.

\bibitem{lee2020biobert}
J.~Lee, W.~Yoon, S.~Kim, D.~Kim, S.~Kim, C.~H. So, and J.~Kang, ``Biobert: a pre-trained biomedical language representation model for biomedical text mining,'' {\em Bioinformatics}, vol.~36, no.~4, pp.~1234--1240, 2020.

\bibitem{peng2019transfer}
Y.~Peng, S.~Yan, and Z.~Lu, ``Transfer learning in biomedical natural language processing: An evaluation of bert and elmo on ten benchmarking datasets,'' in {\em Proceedings of the 2019 Workshop on Biomedical Natural Language Processing (BioNLP 2019)}, pp.~58--65, 2019.

\bibitem{liu2021finbert}
Z.~Liu, D.~Huang, K.~Huang, Z.~Li, and J.~Zhao, ``Finbert: A pre-trained financial language representation model for financial text mining,'' in {\em Proceedings of the twenty-ninth international conference on international joint conferences on artificial intelligence}, pp.~4513--4519, 2021.

\bibitem{kim2023medibiodeberta}
E.~Kim, Y.~Jeong, and M.-s. Choi, ``Medibiodeberta: Biomedical language model with continuous learning and intermediate fine-tuning,'' {\em IEEE Access}, 2023.

\bibitem{10820432}
Z.~R.~K. Rostam and G.~Kertész, ``Fine-tuning large language models for scientific text classification: A comparative study,'' in {\em 2024 IEEE 6th International Symposium on Logistics and Industrial Informatics (LINDI)}, pp.~000233--000238, 2024.

\bibitem{doi:10.7326/ANNALS-24-02189}
``Development of prompt templates for large language model–driven screening in systematic reviews,'' {\em Annals of Internal Medicine}, vol.~178, no.~3, pp.~389--401, 2025.
\newblock PMID: 39993313.

\bibitem{abburi2023generative}
H.~Abburi, M.~Suesserman, N.~Pudota, B.~Veeramani, E.~Bowen, and S.~Bhattacharya, ``Generative ai text classification using ensemble llm approaches,'' {\em arXiv preprint arXiv:2309.07755}, 2023.

\bibitem{10825609}
S.~Z. Ridoy, M.~Shazzad Hossain~Shaon, A.~Cuzzocrea, and M.~S. Akter, ``Enstack: An ensemble stacking framework of large language models for enhanced vulnerability detection in source code,'' in {\em 2024 IEEE International Conference on Big Data (BigData)}, pp.~6356--6364, 2024.

\bibitem{white2020pubmed}
J.~White, ``Pubmed 2.0,'' {\em Medical reference services quarterly}, vol.~39, no.~4, pp.~382--387, 2020.

\bibitem{GooglePricing2026}
{Google}, ``Gemini {API} pricing.'' https://ai.google.dev/gemini-api/docs/pricing, 2026.
\newblock Accessed: Feb. 11, 2026.

\bibitem{PricePerToken2026}
{PricePerToken}, ``Google {API} pricing.'' https://pricepertoken.com/pricing-page/provider/google, 2026.
\newblock Accessed: Feb. 11, 2026.

\end{thebibliography}

\end{document}